\documentclass[10pt,twocolumn,letterpaper]{article}

\usepackage[pagenumbers]{cvpr} 

\usepackage{graphicx}
\usepackage{amsmath}
\usepackage{amssymb}
\usepackage{booktabs}
\usepackage{multirow}
\usepackage{float}
\usepackage[symbol]{footmisc}

\usepackage[pagebackref,breaklinks,colorlinks]{hyperref}

\usepackage[capitalize]{cleveref}
\crefname{section}{Sec.}{Secs.}
\Crefname{section}{Section}{Sections}
\Crefname{table}{Table}{Tables}
\crefname{table}{Tab.}{Tabs.}

\def\cvprPaperID{*****} 
\def\confName{CVPR}
\def\confYear{2022}

\begin{document}

\title{BEVerse: Unified Perception and Prediction in Birds-Eye-View for Vision-Centric Autonomous Driving}


\author{Yunpeng Zhang\textsuperscript{1}, Zheng Zhu\textsuperscript{2}, Wenzhao Zheng\textsuperscript{1}, 
Junjie Huang\textsuperscript{2},\\
Guan Huang\textsuperscript{2},
Jie Zhou\textsuperscript{1},
Jiwen Lu\textsuperscript{1}\thanks{Corresponding author.}\\
\textsuperscript{1}Tsinghua University ~ ~
\textsuperscript{2}PhiGent Robotics
}

\maketitle

\begin{abstract}

In this paper, we present BEVerse, a unified framework for 3D perception and prediction based on multi-camera systems. Unlike existing studies focusing on the improvement of single-task approaches, BEVerse features in producing spatio-temporal Birds-Eye-View (BEV) representations from multi-camera videos and jointly reasoning about multiple tasks for vision-centric autonomous driving. 
Specifically, BEVerse first performs shared feature extraction and lifting to generate 4D BEV representations from multi-timestamp and multi-view images. After the ego-motion alignment, the spatio-temporal encoder is utilized for further feature extraction in BEV. Finally, multiple task decoders are attached for joint reasoning and prediction. Within the decoders, we propose the grid sampler to generate BEV features with different ranges and granularities for different tasks. Also, we design the method of iterative flow for memory-efficient future prediction. 
We show that the temporal information improves 3D object detection and semantic map construction, while the multi-task learning can implicitly benefit motion prediction. 
With extensive experiments on the nuScenes dataset, we show that the multi-task BEVerse outperforms existing single-task methods on 3D object detection, semantic map construction, and motion prediction. Compared with the sequential paradigm, BEVerse also favors in significantly improved efficiency.
The code and trained models will be released\footnote[2]{\url{https://github.com/zhangyp15/BEVerse}}.

\end{abstract}

\section{Introduction}
\label{sec:intro}

\begin{figure}[t]
\centering
\includegraphics[width=1.0\linewidth]{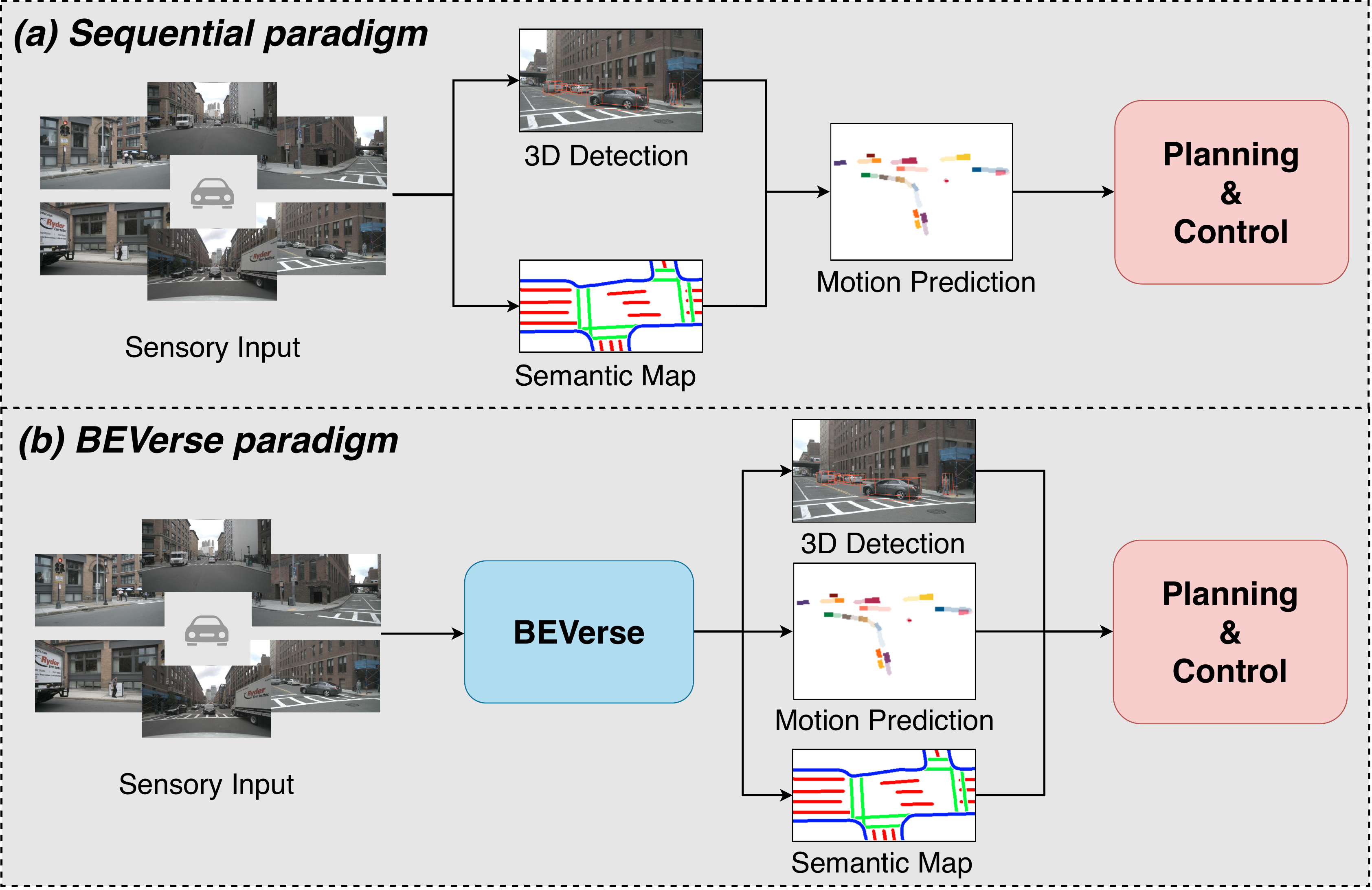}
\caption{The traditional paradigm (a) follows the sequential design, where the perception, prediction, and planning\&control are conducted one by one. Note that the perception includes the detection for dynamic objects and the map construction for static environments. Since the sequential paradigm inevitably suffers from repeated feature extraction and severe error propagation, we propose BEVerse (b) for joint perception and prediction. With shared feature extraction and parallel multi-task inference, BEVerse achieves a better trade-off between performance and efficiency.}
\label{fig:multi_task_motivation}
\vspace{-5mm}
\end{figure}

Modern industrial systems generally follow the principle of problem decomposition. For self-driving vehicles, the whole problem is divided into perception, prediction, and planning\&control.
The task of perception is to perceive the surrounding environment, including the dynamic objects~\cite{pointpillars,centertrack,voxelnet,mono3d,deep3Dbox,pseudo-lidar,fcos3d,detr3d,bevdet} and static streets~\cite{hdmapnet,bevsegformer}. Since the driving scenarios are rapidly changing, the task of prediction~\cite{uncertainty-motion,safety-aware-motion,tpnet,tnt,denseTNT,fiery} is needed to speculate the future movements of recognized obstacles. With the information from perception and prediction, planning~\cite{planning_RL,review-motion-planning,porca_planning,motion-planning-ICRA-2011} tries to determine the accurate and secure driving behavior towards the defined target, while the control system drives the vehicle to carry out the desired behavior.

As shown in Figure~\ref{fig:multi_task_motivation}, the traditional paradigm is to stack these subtasks sequentially, where the output of one subtask is fed into the next as input. The sequential design enables the slicing of single task from the whole system, which creates independent and specific problems for academic researches. However, the propagation of errors can significantly influence the downstream tasks. 
Also, the sequential paradigm can inherently bring extra computational burden due to repeated feature extraction and propagation.
Recent studies~\cite{FAF, PNPNet, JointLP2, motionnet} have been exploring the joint reasoning of perception and prediction for LiDAR-centric autonomous driving systems. These methods have demonstrated that the multi-task paradigm can be more efﬁcient due to shared computations and can also achieve the state-of-the-art performance, benefiting from temporal fusion and joint learning. 
Considering the expensive costs of LiDAR sensors, the vision-centric method relies on multiple surrounding cameras as the input information. Though the cost-effective alternative has been widely studied in the perception~\cite{fcos3d, PGD, detr3d, bevdet, hdmapnet} and prediction~\cite{fiery}, the multi-task paradigm for vision-centric autonomous driving has not been discussed before. 

To this end, we are motivated to propose the first  \textbf{B}irds-\textbf{E}ye-\textbf{V}iew metav\textbf{erse} (\textbf{BEVerse}), for joint perception and prediction in vision-centric autonomous driving. With consecutive frames from multiple surrounding cameras as input, BEVerse constructs 4D feature representations in BEV and jointly reasons about 3D object detection, semantic map construction, and motion prediction. Specifically, we first perform parallel feature extraction for multi-frame and multi-view images and construct the BEV features for each frame with the image-to-BEV view transformer. 
To further exploit temporal information, we transform past BEV features to the present coordinate system with the ego-motions and process the aligned 4D features with the spatio-temporal BEV encoder. 
Finally, the present BEV feature is used as the shared input for multiple task decoders. To satisfy the specific requirement of BEV range and granularity for each task, we propose the grid samplers to crop and transform the input BEV feature before decoding. Also, we observe that existing methods of generating future states for motion prediction can be heavily memory-consuming and prevent multi-task learning. Therefore, we propose the method of iterative flow for efficient future prediction.
To sum up, the proposed BEVerse is a one-stage and multi-task framework which takes 4D sensory input, constructs spatio-temporal BEV feature representations, and jointly reasons about the perception and prediction for autonomous driving. 
We demonstrate the effectiveness of BEVerse with experimental results on the nuScenes~\cite{nuscenes} dataset. For 3D object detection, BEVerse achieves 53.1\% NDS on the test set. For semantic map construction, BEVerse scores 51.7\% mIoU and surpasses the previous arts by 7.1 points. For motion prediction, BEVerse obtains 40.9\% IoU and 36.1\% VPQ, which are 4.2\% IoU and 6.2\% VPQ higher than FIERY~\cite{fiery}. Compared with the sequential paradigm, the multi-task paradigm of BEVerse can also significantly improve the efficiency. 

The main contributions of the paper can be summarized in three aspects: (1) We propose BEVerse, the first framework for unified perception and prediction in Birds-Eye-View with multi-camera autonomous driving systems. (2) We propose the method of iterative flow for efficient future prediction and enabling multi-task learning. 
(3) With one multi-task model, BEVerse achieves the state-of-the-art performance for 3D object detection, semantic map construction, and motion prediction on the nuScenes~\cite{nuscenes} dataset and is more efficient than the sequential paradigm.

\section{Related Work}

\begin{figure*}[t]
\centering
\includegraphics[width=1.0\linewidth]{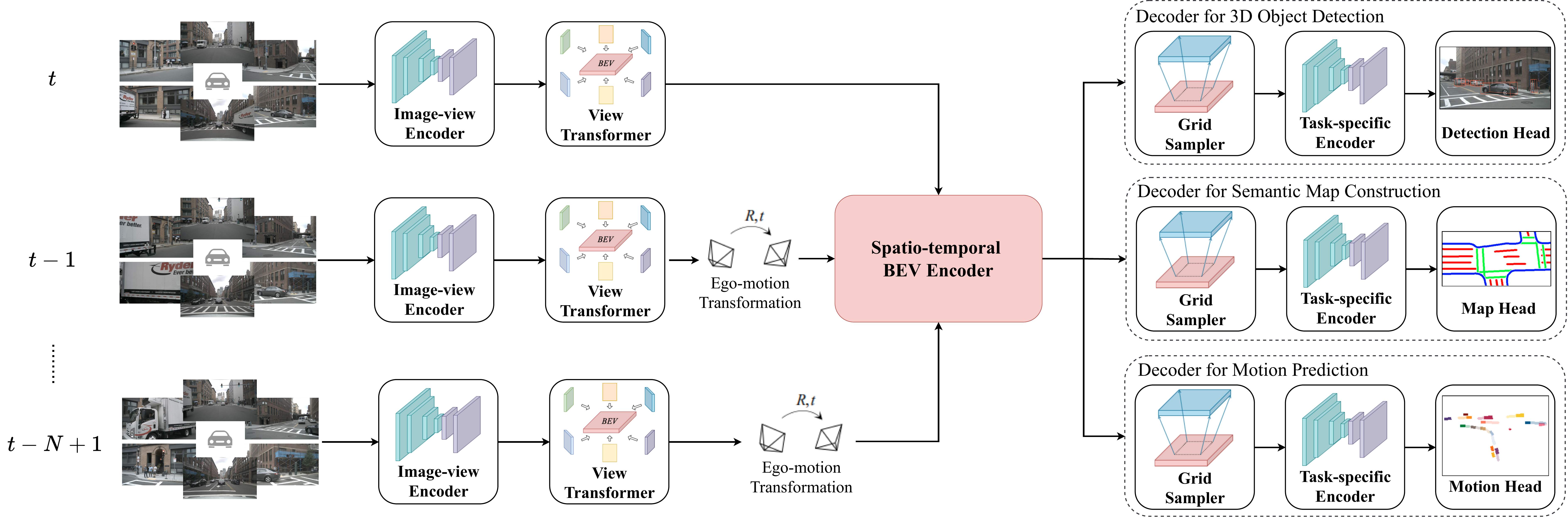}
\caption{The framework of BEVerse. With consecutive frames from surrounding cameras as input, BEVerse first constructs the BEV feature representation for each timestamp. The process includes the image-view feature extraction and the view transformation. Then, the BEV features from past frames are aligned to remove ego-motions and processed by the temporal model. Finally, the well-established BEV feature with both spatial and temporal information is sent to multiple task decoders for joint reasoning of perception and prediction.
}
\vspace{-5mm}
\label{fig:framework}
\end{figure*}

\vspace{-1mm}
\subsection{3D Object Detection}

Considering the utilized sensors, existing methods for 3D object detection can include LiDAR-based, stereo-based, monocular, and multi-modal subcategories. In this paper, we focus on methods with surrounding monocular cameras. Since the release of the KITTI~\cite{kitti} 3D object detection benchmark, plenty of studies~\cite{mono3d,deep3Dbox,FQNet, GS3D, pseudo-lidar, M3D-RPN, monopair, monodle, monoflex, GUPNet, autoshape} have been proposed to improve the monocular 3D object detection. However, the front-view camera in the KITTI dataset cannot perceive the entire environment. The release of nuScenes~\cite{nuscenes}, Waymo Open Dataset~\cite{waymo}, and other modern datasets for autonomous driving has come to provide sufﬁcient training samples for 3D object detection with multiple surrounding cameras. 

FCOS3D~\cite{fcos3d} extends the fully convolutional 2D detector FCOS~\cite{fcos} to monocular 3D object detection by proposing the 3D center-ness and providing a good practice for learning 3D attributes. PGD~\cite{PGD} further improves the instance depth estimation of FCOS3D~\cite{fcos3d} by utilizing geometric relation across objects and probabilistic depth representation. However, both methods separately process the image of each view and merge the outputs with heuristic post-processing. This paradigm fails in taking full advantage of the information from surrounding cameras, especially in the overlap regions, and leads to sub-optimal performance. 
DETR3D~\cite{detr3d} extracts multi-scale features from multiple camera images and uses a set of 3D object queries to index corresponding camera features through geometric back-projection. Finally, these queries generates the bounding box predictions in the DETR-like~\cite{detr} style. 
PETR~\cite{PETR} encodes the information of 3D coordinates as positional encodings to generate 3D position-aware features for DETR-like~\cite{detr} decoding. 
BEVDet~\cite{bevdet} constructs the BEV feature representation from multiple camera images following Lift-Splat-Shoot~\cite{LSS} and proposes the isolated image-view and BEV augmentation strategies to achieve satisfactory performance. In this paper, we basically follow the practices of BEVDet~\cite{bevdet} to build the 3D object detection branch of BEVerse. Also, BEVerse demonstrates the effectiveness of temporal information for 3D object detection.

\vspace{-1mm}
\subsection{Semantic Map Construction}
High-definition maps (HD maps) can provide fine-grained information about the road scenes and play a vital role for autonomous vehicles. Most HD maps are manually annotated, while some studies use SLAM algorithms~\cite{SLAM_2017_iros, icp_pose_graph_SLAM, robust_pose_graph_for_SLAM, robust_pose_graph_EM, lio-sam} to generate HD maps from repeated scanning of the environments with LiDAR sensors. These approaches involve large-scale data collections, long-term iterations, and expensive human annotations. Therefore, building local maps directly from onboard sensors becomes an affordable alternative. Besides, the ability to build semantic maps online is important as a redundant design.
HDMapNet~\cite{hdmapnet} first introduces the problem of local semantic map learning, which aims to construct semantic maps online from the observations of LiDAR sensors and cameras. It also proposes a learning method to build BEV features from sensory input and predicts vectorized map elements.
BEVSegFormer~\cite{bevsegformer} proposes the multi-camera deformable attention to transform image-view features to BEV representations for semantic map construction. 
Different from these single-task approaches, our BEVerse incorporates the semantic map construction as part of the multi-task framework and uses vanilla convolutional layers for segmentation prediction. We also show that the temporal information can improve the learning of semantic maps. 

\vspace{-1mm}
\subsection{Motion Prediction}
The future behaviour of other road agents is important for self-training systems to make safe planning-decisions and plenty of camera-based methods~\cite{patch_to_the_future, predict_parsing_motion, uncertainty-motion, DESIRE, fiery} have been proposed for motion prediction.
\cite{patch_to_the_future} learns to predict the future visual appearances in an unsupervised manner. \cite{predict_parsing_motion} jointly predicts the semantic segmentation and optical flow in the perspective view. However, these methods operate in the image domain and cannot offer enough information for 3D planning. 
Recent methods~\cite{uncertainty-motion, DESIRE} produce raster images of semantic maps and states of road agents as the input for motion prediction. These frameworks are based on the results of detection and require the presence of HD maps. 
FIERY~\cite{fiery} proposes the first framework for BEV motion prediction directly from the videos of surrounding cameras. The concurrent StretchBEV~\cite{stretchbev} further propose to sample the latent variables at each timestamp and predict residual changes for producing future states. 
Similar to FIERY~\cite{fiery}, our method also takes the raw sensory input for joint perception and prediction in BEV coordinates. To reduce the memory consumption of FIERY~\cite{fiery} and enable the inference of multiple tasks, we proposes the iterative flow for the efficient generation of future states.

\subsection{Multi-Task Learning}
Multi-task learning~\cite{overview_mtl_2017, overview_mtl_2018} aims to jointly solve multiple related tasks with shared network structures and mutual promotions across tasks. General studies for multi-task learning focus on how to design the shared structures~\cite{multilinear_relation, cross-stitch, nddr-cnn, latent_mtl, learning_to_branch_mtl} and how to balance the multi-task optimization~\cite{what_uncertainty, gradnorm, gradient_surgery, gradient_sign_dropout}.
For autonomous driving, multi-task frameworks~\cite{FAF, PNPNet, motionnet, JointLP2} have been widely discussed for LiDAR-centric systems. FAFNet~\cite{FAF} proposes a holistic model that jointly reasons about detection, prediction, and tracking. MotionNet~\cite{motionnet} proposes a hierarchical spatio-temporal pyramid network to encode BEV features from a sequence of LiDAR sweeps. Then, it performs joint perception and motion prediction without using bounding boxes.
In this paper, we propose the first multi-task approach for unified perception and prediction in BEV for camera-centric systems. 

\vspace{-1mm}
\section{Approach}

BEVerse takes $M$ surrounding camera images from $N$ timestamps and the corresponding ego-motions and camera parameters as input. With multi-task inference, the outputs includes the 3D bounding boxes and semantic map for the present frame, and the future instance segmentation and motion for the following $T$ frames. 
As shown in Figure~\ref{fig:framework}, the proposed framework BEVerse consists of four sub-modules that are sequentially applied, including the image-view encoder, the view transformer, the spatio-temporal BEV encoder, and the multi-task decoders. We elaborate on these modules in the following subsections.

\subsection{Image-view Encoder}
\label{sec:img-view-encoder}

Assuming each input image is of shape $H \times W \times 3$, the image-view encoder performs the feature extraction in the perspective view, which is shared across different cameras and timestamps. Since the complex driving scenario includes elements of various sizes, multi-scale features are exploited within the encoder. Specifically, we use the recently proposed SwinTransformer~\cite{swin_transformer} as the backbone network to create multi-level features $C_2$, $C_3$, $C_4$, $C_5$, where $C_i$ denotes the feature of spatial size $\frac{H}{2^i} \times \frac{W}{2^i}$. To enable a efficient fusion, we follow BEVDet~\cite{bevdet} to upsample $C_5$ by 2$\times$ resolution and concatenate it with $C_4$. Then two convolutional layers are carried out to form the output feature map $F$ with shape $\frac{H}{16} \times \frac{W}{16} \times C$, where $C$ refers to the feature channel. 
\subsection{View Transformer}

With the requirement of learning 3D temporal information and predicting multiple BEV tasks, the image-to-BEV transformation is one essential procedure in enabling BEVerse. For each time stamp, the view transformer takes the multi-view features $F \in \mathbb{R}^{M \times H' \times W' \times C}$ and outputs the BEV feature $G \in \mathbb{R}^{X \times Y \times C}$ covering the entire surroundings, where $X$ and $Y$ are the sizes of defined BEV grids. 
Specifically, we adopt the method proposed in Lift-Splat-Shoot~\cite{LSS} for view transformation. The multi-view features $F$ are processed by one 1$\times$1 convolution to predict the categorical depth distribution $F \in \mathbb{R}^{M \times H' \times W' \times D}$, where $D$ is the number of predefined depth bins. Then each pixel on the feature map is lifted to $D$ points with these depths and camera matrices. The generated dense point cloud with $M \times H' \times W' \times D$ points is then processed with pillar pooling~\cite{pointpillars} to create the BEV feature representation $G \in \mathbb{R}^{X \times Y \times C}$. 

\subsection{Spatio-temporal BEV Encoder}

\begin{figure}[t]
\centering
\begin{subfigure}{1.0\linewidth}
\includegraphics[width=\linewidth]{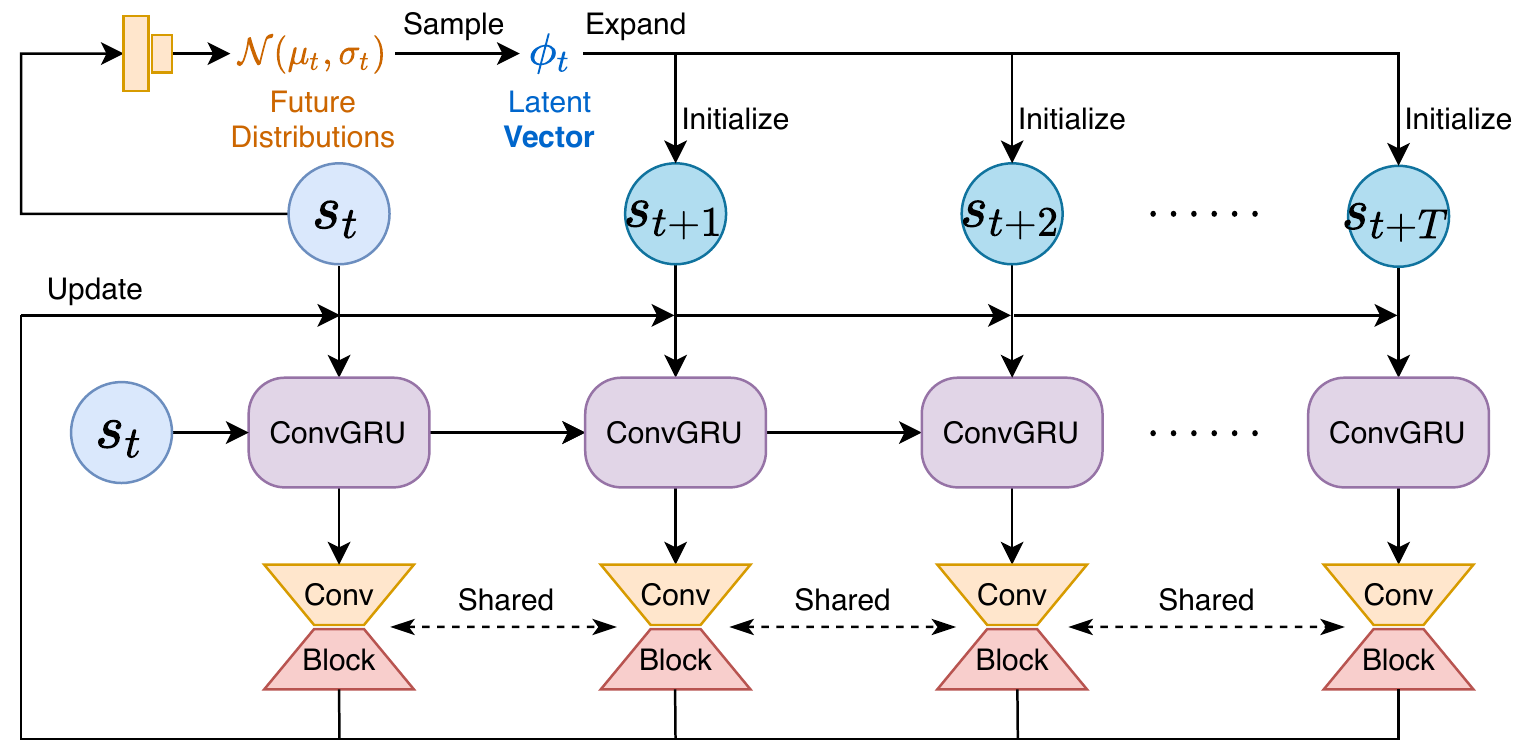}
\caption{The future prediction of FIERY~\cite{fiery}.}
\end{subfigure}
\begin{subfigure}{1.0\linewidth}
\includegraphics[width=\linewidth]{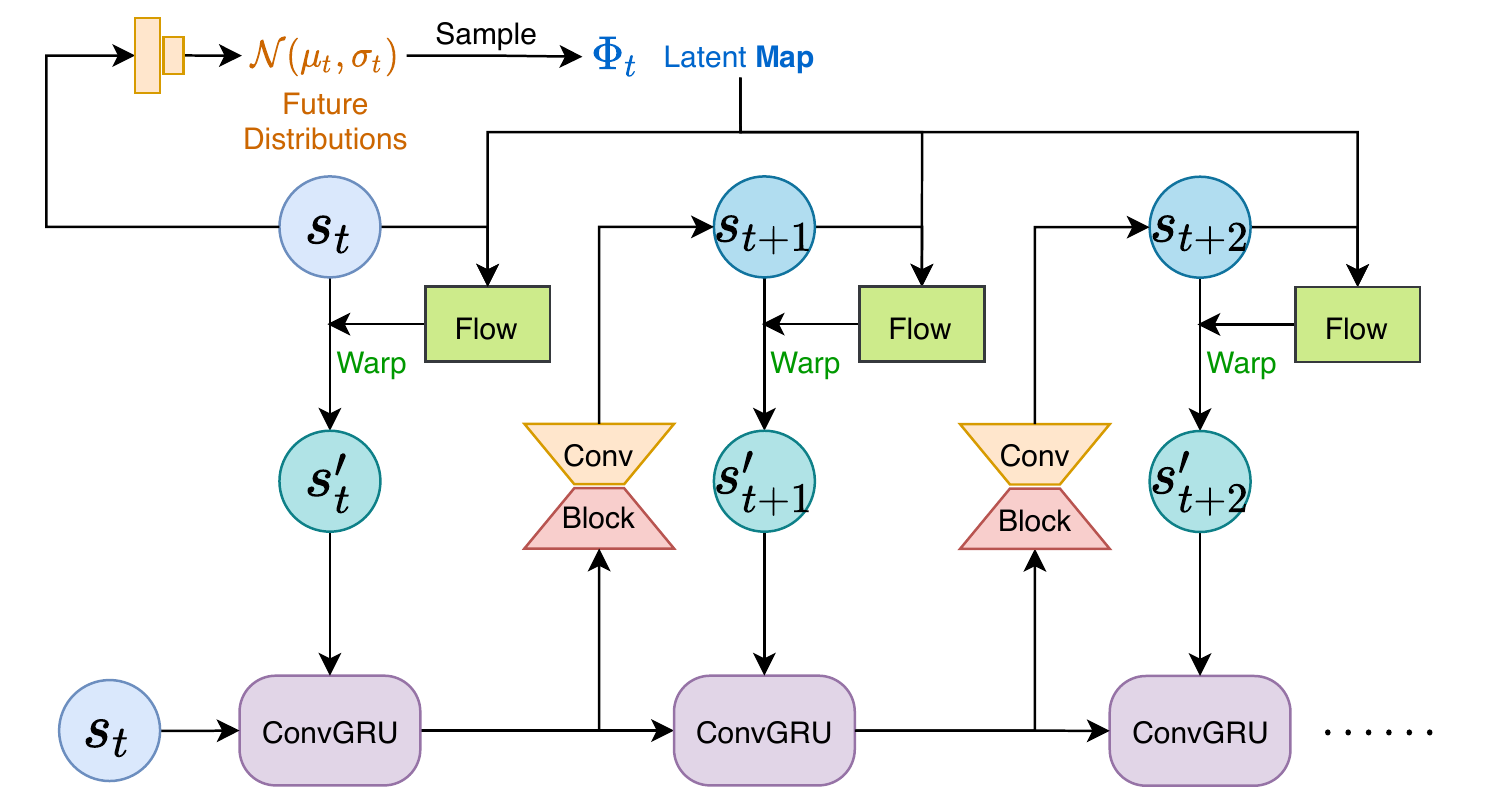}
\caption{The future prediction of the proposed iterative flow.}
\end{subfigure}
\caption{The qualitative comparison between the methods for future prediction from FIERY~\cite{fiery} and the proposed iterative flow.}
\vspace{-3mm}
\label{fig:future_compare}
\end{figure}

After the view transformation, we acquire the BEV features $G \in \mathbb{R}^{N \times X \times Y \times C}$ from $N$ past timestamps. Since movements of the ego-vehicle can cause the coordinate misalignment of different timestamps, we first warp past features to the present reference frame with known ego-motions. The aligned 4D tensor is then processed with a spatio-temporal BEV encoder to further extract both spatial and temporal information. Following FIERY~\cite{fiery}, the BEV encoder is formulated as the stack of temporal blocks. Each block mainly consists of 3D convolutions, global pooling operations, and in-between feature compression layers. Finally, we get the spatio-temporal BEV feature of the present frame $G_p \in \mathbb{R}^{X \times Y \times C_o}$ as the input for multi-task decoders.

\begin{table*}[t]
\centering
\caption{Ablation on the incorporation of temporal information for 3D object detection.}
\resizebox{1.0\linewidth}{!}{
\begin{tabular}{c|c|cc|ccccccc}
\toprule
Method & Temporal & \#param. & GFLOPs & NDS$\uparrow$ & mAP$\uparrow$ & mATE$\downarrow$  & mASE$\downarrow$   & mAOE$\downarrow$  & mAVE$\downarrow$  &  mAAE$\downarrow$\\
\midrule
\multirow{2}{*}{BEVerse-Det} & & 55.7M & 215.3 & 0.392 & 0.302 & 0.687 & 0.278 & 0.566 & 0.838 & 0.219\\
& \checkmark & 55.8M & 220.0 & \textbf{0.474} & \textbf{0.327} & \textbf{0.675} & \textbf{0.274} & \textbf{0.430} & \textbf{0.324} &	\textbf{0.192}\\
\bottomrule
\end{tabular}}
\label{tab:ablation_temporal_det}
\end{table*}

\begin{table*}[t]
\centering
\caption{Ablation on the incorporation of temporal information for semantic map construction.}
\begin{tabular}{c|c|cc|cccc}
\toprule
Method & Temporal & \#param. & GFLOPs & Divider & Ped Cross & Boundary & mIoU\\
\midrule
\multirow{2}{*}{BEVerse-Map} & & 54.8M & 429.7 & 53.9 & 41.0 & 54.5 & 49.8\\
& \checkmark & 54.9M & 434.4 & \textbf{56.1} & \textbf{44.9} & \textbf{58.7} & \textbf{53.2}\\
\bottomrule
\end{tabular}
\label{tab:ablation_temporal_map}
\end{table*}

\begin{table*}[!ht]
\centering
\caption{Ablation on the approach for future prediction. ``GPU Memory'' is the training memory with batch size as 1. *: we use the module for future prediction from FIERY~\cite{fiery} to replace the corresponding module in our model for comparison. }
\begin{tabular}{c|ccc|cc|cc}
\toprule
\multirow{2}{*}{Method} & \multirow{2}{*}{GPU Memory} & \multirow{2}{*}{GFLOPs} & \multirow{2}{*}{\#param.}
& \multicolumn{2}{c|}{IoU}
& \multicolumn{2}{c}{VPQ}\\ 
& & & & Short & Long & Short & Long \\
\midrule
FIERY$^*$~\cite{fiery} & $\sim$22G & 2503.2 & 68.3M & 59.4 & 36.7 & 50.2 & 29.9 \\
\midrule
Iterative baseline & $\sim$11G & 1439.0 & 61.7M & 58.5 & 36.4 & 48.2 & 29.4 \\
+ Flow warp & $\sim$12G & 1558.7 & 62.4M & 59.5 & 36.7 & 50.9 & 31.0 \\
BEVerse-Motion & $\sim$12G & 1558.7 & 62.4M & \textbf{59.7} & \textbf{37.5} & \textbf{51.0} & \textbf{32.0} \\
\bottomrule
\end{tabular}
\label{tab:ablation_future_pred}
\end{table*}

\subsection{Task Decoders}
With the BEV feature $G_p$ as input, BEVerse employs independent and parallel decoders for the joint reasoning of perception and prediction. Each task decoder includes the grid sampler, the task encoder, and the task head. In the following paragraphs, we first introduce the sampler and encoder, which share similar structures across tasks. Then we detail the design of three different task heads.

\noindent\textbf{Grid sampler.} 
Since different tasks may require specific range and granularity, the spatial scope and resolution of the input BEV feature cannot directly serve for decoding. For example, the learning of semantic map requires fine-grained features because the traffic lines are quite narrow in the 3D space.
Therefore, the grid sampler is proposed to crop task-specific regions and transformed to ideal resolution through bi-linear interpolation. In our experiments, we set the base BEV grid to be large and coarse for efficiency. 

\noindent\textbf{Task encoder.} 
After the feature sampling, a lightweight task encoder is applied to encode task-specific features in the corresponding BEV grids. Following BEVDet~\cite{bevdet}, we utilize the basic block in ResNet~\cite{resnet} to build the backbone and combine the multi-scale features similar to the image-view encoder in~\ref{sec:img-view-encoder}. The output feature is upsampled to the input resolution and sent to the task head. 

\noindent\textbf{Head for 3D object detection.} 
Since the BEV feature representation has been constructed from multiple camera videos, the dimension gap between monocular and LiDAR-based methods has been bridged. Therefore, the state-of-the-art detection heads designed for LiDAR can be directly adopted without modification. In this paper, we simply use the first stage of CenterPoint~\cite{centertrack} as the head for 3D object detection.

\noindent\textbf{Head for semantic map construction.}
We construct the head for semantic map construction with two vanilla convolutional layers with BatchNorm~\cite{batchnorm} and ReLU. The output channel is the number of classes $C_\text{map}$ in semantic maps.

\noindent\textbf{Head for motion prediction.} Different from the above heads which only care about the present frame, motion prediction also requires the forecast of future states. As illustrated in Figure~\ref{fig:future_compare}, FIERY~\cite{fiery} first predicts the parameters of future Gaussian distributions and samples a latent vector $\phi_t \in \mathbb{R}^{L}$, where $L$ is the latent dimension. The sampled $\phi_t$ is spatially expanded to shape $\mathbb{R}^{X_\text{motion} \times Y_\text{motion} \times L}$ and used to initialize the future states. Then, the convolutional gated recurrent unit network and the bottleneck block~\cite{resnet} are repeated applied for generating the future states $\{s_{t+1}, s_{t+2}, \cdots, s_{t+T}\}$. Two important factors influence the effectiveness of FIERY's~\cite{fiery} prediction module: (1) The sampled global latent vector $\phi_t$ is shared for each BEV pixel and cannot represent the uncertainties of many different agents. (2) Initializing future states from only sampled latent vectors can increase the difficulty of prediction. To this end, we propose the iterative flow for future prediction. Different from FIERY~\cite{fiery}, we directly predict and sample a latent map $\mathbb{R}^{X_\text{motion} \times Y_\text{motion} \times L}$ so that the uncertainties of different objects can be separated. Also, the state of next timestamp is generated by warping the current state with predicted flows, which naturally adapts to the problem of motion prediction and simplifies the learning process. After the future states are generated, we use the same output heads like FIERY~\cite{fiery} to predict future instance segmentation and motion.

\begin{table*}[t]
\centering
\caption{Ablation on the influence of multi-task learning.}
\resizebox{1.0\linewidth}{!}{
\begin{tabular}{c|ccc|cc|cccc|cc}
\toprule
\multirow{2}{*}{Method} & \multirow{2}{*}{\#param.} & \multirow{2}{*}{GFLOPs} & \multirow{2}{*}{FPS} 
& \multicolumn{2}{c|}{Detection}
& \multicolumn{4}{c|}{Semantic map}
& \multicolumn{2}{c}{Motion} \\ 
& & & & NDS & mAP & Divider & Ped Cross & Boundary & mIoU & IoU & VPQ\\
\midrule
BEVerse-Det & 55.8M & 220.0 & 12.6 & 0.474 & 0.327 & - & - & - & - & - & - \\
BEVerse-Map & 54.9M & 434.4 & 14.2 & - & - & 56.1 & 44.9 & 58.7 & 53.2 & - & -\\
BEVerse-Motion & 62.4M & 1558.7 & 5.6 & - & - & - & - & - & - & 37.5 & 32.0\\
Sequential & 173.1M & 2213.1 & 3.0 & 0.474 & 0.327 & 56.1 & 44.9 & 58.7 & 53.2 & 37.5 & 32.0\\
\midrule
BEVerse & 102.5M & 1921.9 & 4.4 & 0.466 & 0.321 & 53.2 & 39.0 & 53.9 & 48.7 & 38.7 & 33.3\\
\bottomrule
\end{tabular}}
\label{tab:ablation_multi_task}
\end{table*}

\begin{table*}[t]
\centering
\caption{Ablation on the strategies for data-augmentation.}
\begin{tabular}{cc|cc|cccc|cc}
\toprule
\multicolumn{2}{c|}{Augmentation}
& \multicolumn{2}{c|}{Detection}
& \multicolumn{4}{c|}{Semantic map}
& \multicolumn{2}{c}{Motion} \\ 
Image-view & BEV & NDS & mAP & Divider & Ped Cross & Boundary & mIoU & IoU & VPQ\\
\midrule
& & 0.323 & 0.152 & 49.2 & 34.9 & 51.7 & 45.3 & 31.2 & 27.2\\
\checkmark & & 0.383 & 0.203 & 51.6 & 37.8 & 53.8 & 47.8 & 32.9 & 28.8\\
& \checkmark & 0.393 & 0.208 & 47.2 & 34.6 & 49.2 & 43.7 & 32.8 & 27.8\\
\checkmark & \checkmark & \textbf{0.466} & \textbf{0.321} & \textbf{53.2} & \textbf{39.0} & \textbf{53.9} & \textbf{48.7} & \textbf{38.7} & \textbf{33.3}\\
\bottomrule
\end{tabular}
\label{tab:ablation_augmentation}
\end{table*}

\section{Experiments}

\subsection{Dataset}
To demonstrate the effectiveness of BEVerse, we conduct comprehensive experiments on the nuScenes~\cite{nuscenes} dataset, which is widely used as the benchmark for both perception and prediction in autonomous driving. The nuScenes~\cite{nuscenes} dataset includes 1000  driving video clips collected in Boston and Singapore. Each clip is 20 seconds long and annotated with 3D bounding boxes at 2Hz, generating up to 40k key-frames and 1.4M object bounding boxes. All clips are officially divided into the splits for training, validation, and test with 700, 150, and 150 clips. 
For vision-centric methods, the provided sensory input includes six surrounding cameras, intrinsic/extrinsic matrices, and ego-motions. 

\subsection{Metrics}

\noindent\textbf{3D object detection.} 
The official evaluation metrics for 3D object detection include the Average Precision metric and a set of True Positive metrics. The former is formulated as the mean Average Precision (mAP), which measures the area under the precision-recall curve. To match predictions and ground-truth objects, nuScenes requires the BEV center distance to be less than a given threshold. The average is performed over four thresholds and ten classes. The True Positive metrics include Average Translation Error (ATE), Average Scale Error (ASE), Average Orientation Error (AOE), Average Velocity Error (AVE), and Average Attribute Error (AAE). Finally, the weighted summation of mAP and these True Positive metrics is computed to form the nuScenes detection score (NDS) as the ranking metric. 

\noindent\textbf{Semantic map construction.} Following \cite{hdmapnet}, the semantic classes for map construction include lane dividers, pedestrian crossings, and lane boundaries. For quantitative evaluation, we compute the intersection-over-union (IoU) for each class between the predicted and ground-truth maps. The mean IoU (mIoU) is computed as the ranking metric. 

\noindent\textbf{Motion prediction.} 
Following \cite{fiery}, we use IoU and VPQ (Future Video Panoptic Quality) to evaluate the performance of motion prediction. The IoU measures the segmentation of objects at present and future frames, while the VPQ evaluates both the recognition and segmentation quality of the predicted trajectories. The VPQ is computed as~\eqref{equ:metric_vpq}:
\begin{equation}
    \text{VPQ} = \sum_{t=0}^T \frac{\sum_{(p_t,q_t) \in TP_t} \text{IoU}(p_t,q_t)}{|TP_t| + \frac{1}{2}|FP_t| + \frac{1}{2}|FN_t|}
    \label{equ:metric_vpq}
\end{equation}
where $TP_t$, $FP_t$, and $FN_t$ are the set of true positives, false positives, and false negatives at timestamp $t$. Similar to~\cite{fiery}, we also compute both metrics under two different spatial settings: $30 \mathrm{m} \times 30 \mathrm{m}$ (Short) and $100 \mathrm{m} \times 100 \mathrm{m}$ (Long) around the ego vehicle. 

\begin{table*}[t]
\centering
\caption{Comparison with the state-of-the-art methods for 3D object detection on the nuScenes~\cite{nuscenes} validation set. 
$\dag$: the feature encoder is initialized from a FCOS3D~\cite{fcos3d} checkpoint. 
$\S$: with test-time augmentation. 
$\#$: with model ensemble.}
\resizebox{\linewidth}{!}{
\begin{tabular}{l|cc|ccccccc}
\toprule
Methods  &Image Size & Modality & \textbf{NDS}$\uparrow$ & mAP$\uparrow$ & mATE$\downarrow$  & mASE$\downarrow$   & mAOE$\downarrow$  & mAVE$\downarrow$  &  mAAE$\downarrow$\\
\midrule
CenterFusion~\cite{centerfusion} &800$\times$450 & Camera \& Radar& 0.453 & 0.332   & 0.649             & 0.263            & 0.535             & 0.540             & 0.142 \\
VoxelNet~\cite{voxelnet} & - & LiDAR & 0.648 & 0.563 & 0.292 & 0.253 & 0.316 & 0.287 & 0.191\\
PointPillar~\cite{pointpillars} & - & LiDAR & 0.597 & 0.487 & 0.315 & 0.260 & 0.368 & 0.323 & 0.203\\
\hline
CenterNet~\cite{centernet}  & - & Camera & 0.328 & 0.306 & 0.716 & 0.264 & 0.609 & 1.426 & 0.658\\
FCOS3D~\cite{fcos3d} & 1600$\times$900 & Camera & 0.372 & 0.295 & 0.806 & 0.268 & 0.511 & 1.315 & 0.170\\
DETR3D~\cite{detr3d} & 1600$\times$900 & Camera & 0.374 & 0.303 & 0.860 & 0.278 & 0.437 & 0.967 & 0.235\\
PGD~\cite{PGD} & 1600$\times$900 & Camera & 0.409 & 0.335 & 0.732 & 0.263 & 0.423 & 1.285 & 0.172\\
PETR-R50~\cite{PETR} & 1056$\times$384  & Camera & 0.381 & 0.313 & 0.768 & 0.278 & 0.564 & 0.923 & 0.225\\
PETR-R101~\cite{PETR} & 1408$\times$512 & Camera & 0.421 & 0.357 & 0.710 & 0.270 & 0.490 & 0.885 & 0.224\\
PETR-Tiny~\cite{PETR} & 1408$\times$512 & Camera & 0.431 & 0.361 & 0.732 & 0.273 & 0.497 & 0.808 & 0.185\\
BEVDet-Tiny~\cite{bevdet} & 704$\times$256 & Camera & 0.392 & 0.312 & 0.691 & 0.272 & 0.523 & 0.909 & 0.247\\
BEVDet-Base~\cite{bevdet} & 1600$\times$640 & Camera & 0.472 & 0.393 & 0.608 & 0.259 & 0.366 & 0.822 & 0.191\\
BEVerse-Tiny & 704$\times$256 & Camera & 0.466 & 0.321  & 0.681 & 0.278 & 0.466 & 0.328 & 0.190\\
BEVerse-Small & 1408$\times$512 & Camera & \textbf{0.495} & 0.352 & 0.625	& 0.270 & 0.401 & 0.330 & 0.183\\
\midrule
FCOS3D$\dag\S\#$~\cite{fcos3d}&1600$\times$900 & Camera & 0.415 & 0.343 & 0.725 & 0.263 & 0.422 & 1.292 & 0.153\\
DETR3D$\dag$~\cite{detr3d}  &1600$\times$900 & Camera & 0.434 & 0.349 & 0.716 & 0.268 & 0.379 & 0.842 & 0.200\\
PGD$\dag\S$ \cite{PGD} & 1600$\times$900 & Camera & 0.428 & 0.369 & 0.683 & 0.260 & 0.439 & 1.268 & 0.185\\
PETR-R101$\dag$ \cite{PETR} & 1600$\times$900 & Camera & 0.442 & 0.370 & 0.711 & 0.267 & 0.383 & 0.865 & 0.201\\
BEVDet-Base$\S$~\cite{bevdet} & 1600$\times$640 & Camera & 0.477 & 0.397 & 0.595 & 0.257 & 0.355 & 0.818 & 0.188 \\
BEVerse-Small$\S$ & 1408$\times$512 & Camera & \textbf{0.497} & 0.352 & 0.618 & 0.266 & 0.394 & 0.326 & 0.183\\
\bottomrule
\end{tabular}%
}
\label{tab:nus_det_val}%
\vspace{-5mm}
\end{table*}%

\begin{table}
\begin{center}
\caption{Comparison with the state-of-the-art methods for semantic map construction on the nuScenes~\cite{nuscenes} validation set. $\star$: results reported in HDMapNet~\cite{hdmapnet}.}
\resizebox{\linewidth}{!}{
\begin{tabular}{l|cccc}
\toprule
\multirow{2}{*}{Method} & \multicolumn{4}{c}{Semantic Map IoU}\\
 & Divider & Ped Cross & Boundary & mIoU\\
\midrule
IPM (B)$^\star$ & 25.5 & 12.1 & 27.1 & 21.6 \\
IPM (BC)$^\star$ & 38.6 & 19.3 & 39.3 & 32.4 \\
LSS$^\star$~\cite{LSS} & 38.3 & 14.9 & 39.3 & 30.8 \\
VPN$^\star$~\cite{VPN} & 36.5  & 15.8  & 35.6 & 29.3 \\
HDMapNet$^\star$~\cite{hdmapnet} & 40.6 & 18.7 & 39.5 & 32.9 \\
BEVSegFormer~\cite{bevsegformer} & 51.1 & 32.6 & 50.0 & 44.6 \\
\hline
BEVerse-Tiny & 53.2 & 39.0 & 53.9 & 48.7 \\
BEVerse-Small & \textbf{53.9} & \textbf{44.7} & \textbf{56.4} & \textbf{51.7} \\
\bottomrule
\end{tabular}}
\label{tab:nus_map_val}
\vspace{-3mm}
\end{center}
\end{table}

\subsection{Experimental Settings}
We construct two versions of BEVerse, namely BEVerse-Tiny and BEVerse-Small, for different trade-offs between performance and efficiency. BEVerse-Tiny uses the Swin-T~\cite{swin_transformer} as the backbone and scales the input image to 704$\times$256, while BEVerse-Small uses the stronger Swin-S~\cite{transformer} and scales the image to 1408$\times$512. Note that the raw resolution is 1600$\times$900 in the nuScenes~\cite{nuscenes} dataset. Following the settings of FIERY~\cite{fiery}, BEVerse takes the past three frames (including the present), perceives the present environment, and predicts the instance motion in the upcoming four frames (2.0s for nuScenes). 
We construct the BEV coordinates based on the ego-vehicle system in nuScenes~\cite{nuscenes}. For 3D object detection, we define the BEV ranges are [-51.2m, 51.2m] for both $X$-axis and $Y$-axis, with the interval as 0.8m. For semantic map construction, the ranges are [-30.0m, 30.0m] for $X$-axis and [-15.0m, 15.0m] for $Y$-axis, with the interval as 0.15m. For motion prediction, the ranges are [-50.0m, 50.0m] for both $X$-axis and $Y$-axis, with the interval as 0.5m. The BEV grids of the view transformer follow the settings for detection. 

For implementations of the model architecture, the output channel of the image-view encoder is 512 and is further reduced to 64 during the view transformation. After the temporal model and task-specific encoder, the feature channel increases to 256 for decoding. For the loss weights within each task, we follow the settings of CenterPoint~\cite{centertrack} and FIERY~\cite{fiery}. To balance the learning of multiple tasks, we set the weights for detection, map, and motion as [1.0, 10.0, 1.0]. Unless specified, all reported results are produced with our multi-task framework. 

For training, the AdamW~\cite{adamw} optimizer is utilized, with initial learning rate as 2e-4, weight decay as 0.01, and gradient clip as 35. The model is trained for 20 epochs with CBGS~\cite{cbgs}. For learning schedule, we apply the one-cycle policy~\cite{second} with the peak learning rate as 1e-3. We train the model with a batch size of 64/32 on 32 NVIDIA GeForce RTX 3090 GPUs for BEVerse-Tiny/Small. The backbone is pretrained on ImageNet~\cite{krizhevsky2012imagenet} and other parameters are randomly initialized.
For inference, the scale-NMS and acceleration trick proposed in BEVDet~\cite{bevdet} is adopted. 

For the augmentation strategy, we strictly follow the settings of BEVDet~\cite{bevdet} to perform both the image-view and BEV augmentations. The image-view operations include random scaling, rotation, and flip of the input images. The BEV augmentation include similar operations, but applied to the BEV representations and corresponding learning targets. We apply the same augmentation operations for every past frame for consistency.

\begin{table*}[t]
\centering
\caption{Comparison with the state-of-the-art methods for 3D object detection on the nuScenes~\cite{nuscenes} test set. $\ddag$: the feature encoder is pretrained with depth estimation on the DDAD~\cite{ddad} dataset.}
\resizebox{0.9\linewidth}{!}{
\begin{tabular}{l|c|ccccccc}
\toprule
Methods & Modality & NDS$\uparrow$ & mAP$\uparrow$ & mATE$\downarrow$   & mASE$\downarrow$  & mAOE$\downarrow$  & mAVE$\downarrow$  &  mAAE$\downarrow$\\
\midrule
PointPillars(Light)~\cite{pointpillars}& LiDAR & 0.453 & 0.305 & 0.517 & 0.290 & 0.500 & 0.316 & 0.368\\
CenterFusion~\cite{centerfusion}& Camera \& Radar & 0.449 & 0.326 & 0.631 & 0.261 & 0.516 & 0.614 & 0.115 \\
CenterPoint~\cite{centertrack} & Camera \& LiDAR \& Radar & 0.714 & 0.671 & 0.249 & 0.236 & 0.350 & 0.250 & 0.136\\
\midrule
MonoDIS~\cite{disentangling} & Camera & 0.384 & 0.304 & 0.738 & 0.263 & 0.546 & 1.553 & 0.134\\
CenterNet~\cite{centernet} & Camera & 0.400 & 0.338 & 0.658 & 0.255 & 0.629 & 1.629 & 0.142\\
FCOS3D~\cite{fcos3d} & Camera & 0.428 & 0.358 & 0.690 & 0.249 & 0.452 & 1.434 & 0.124\\
PGD~\cite{PGD} & Camera & 0.448 & 0.386 & 0.626  & 0.245 & 0.451 & 1.509 & 0.127\\
PETR~\cite{PETR} & Camera & 0.481 & 0.434 & 0.641 & 0.248 & 0.437 & 0.894 & 0.143\\
BEVDet~\cite{bevdet} & Camera & 0.482 & 0.422 & 0.529  & 0.236 & 0.395 & 0.979 & 0.152\\
BEVerse & Camera & \textbf{0.531} & 0.393 & 0.541 & 0.247 & 0.394 & 0.345 & 0.129\\
\midrule
DD3D$\ddag$~\cite{dd3d} & Camera & 0.477 & 0.418 & 0.572 & 0.249 & 0.368 & 1.014 & 0.124\\
DETR3D$\ddag$~\cite{detr3d} & Camera & 0.479 & 0.386 & 0.626 & 0.245 & 0.394 & 0.845 & 0.133\\
PETR$\ddag$~\cite{PETR} & Camera & 0.504 & 0.441 & 0.593  & 0.249 & 0.383 & 0.808 & 0.132\\
\bottomrule
\end{tabular}}
\label{tab:nus_det_test}
\vspace{-5mm}
\end{table*}%

\begin{table}[t]
\centering
\caption{Comparison with the state-of-the-art methods for future instance segmentation on the nuScenes~\cite{nuscenes} validation set. $\star$: results reported in FIERY~\cite{fiery}.}
\resizebox{\linewidth}{!}{
\begin{tabular}{l|cc|cc}
\toprule
\multirow{2}{*}{Method}
& \multicolumn{2}{c|}{IoU}
& \multicolumn{2}{c}{VPQ}\\ 
& Short & Long & Short & Long \\
\midrule
Static model$^\star$ & 47.9 & 30.3 & 43.1 & 24.5 \\
Extrapolation model$^\star$ & 49.2 & 30.8 & 43.8 & 24.9\\
FIERY~\cite{fiery}$^\star$ & 59.4 & 36.7 & 50.2 & 29.9 \\
\midrule
BEVerse-Tiny & 60.3 & 38.7 & 52.2 & 33.3 \\
BEVerse-Small & \textbf{61.4} & \textbf{40.9} & \textbf{54.3} & \textbf{36.1} \\
\bottomrule
\end{tabular}}
\label{tab:nus_motion_val}
\vspace{-3mm}
\end{table}

\subsection{Ablation Studies}

For all experiments in the ablation studies, we follow the settings of the backbone and input image sizes for BEVerse-Tiny. Also, we use BEVerse-Det, BEVerse-Map, and BEVerse-Motion to represent the single-task variants of BEVerse.

\noindent\textbf{Temporal information.} In Table~\ref{tab:ablation_temporal_det} and~\ref{tab:ablation_temporal_map}, we analyze the effectiveness of temporal information. For 3D object detection, using the information from past frames can improve mAP by 2.5 points possibly because some occluded objects can be detected with past clues. Also, the estimation of object velocity and orientation is much more accurate with temporal information. For semantic map construction, the incorporation of temporal information also brings a significant improvement.

\noindent\textbf{Future prediction.} 
In Table~\ref{tab:ablation_future_pred}, we compare different methods for generating the future states. Since the future prediction module of FIERY~\cite{fiery} involves repeated structures of ConvGRU and Bottleneck blocks~\cite{resnet}, the module requires too much GPU memory and prevents multi-task learning. With the motivation that future states should be iteratively generated from the last timestamp, the lightweight iterative baseline is proposed and can already achieve comparable performance with FIERY~\cite{fiery} with half memory. Similar to the motion of objects, we introduce the warp with predicted flow and significantly improve the VPQ. Since the spatially shared distribution vector of FIERY~\cite{fiery} cannot well describe the uncertainties of all objects, we directly utilize 2D spatial distribution maps and further improves the performance.

\begin{table}[t]
\centering
\caption{The comparison of the performance and efficiency between the sequential paradigm and the proposed BEVerse with multi-task reasoning. $\dag$: the feature encoder is initialized from a FCOS3D~\cite{fcos3d} checkpoint.}
\resizebox{\linewidth}{!}{
\begin{tabular}{l|c|cc|c|cc}
\toprule
\multirow{2}{*}{Method} & \multirow{2}{*}{FPS} & \multicolumn{2}{c|}{Detection} & Map & \multicolumn{2}{c}{Motion}\\
& & mAP & NDS & mIoU & IoU & VPQ\\ 
\midrule
DETR3D\dag~\cite{detr3d} & 2.0 & 0.349 & 0.434 & - & - & - \\
HDMapNet~\cite{hdmapnet} & 38.7 & - & - & 32.9 & - & - \\
FIERY~\cite{fiery} & 2.1 & - & - & - & 36.7 & 29.9\\
Sequential & 1.0 & 0.349 & 0.434 & 32.9 & 36.7 & 29.9\\
\midrule
BEVerse-Tiny & 4.4 & 0.321 & 0.466 & 48.7 & 38.7 & 33.3\\
BEVerse-Small & 2.6 & \textbf{0.352} & \textbf{0.495} & \textbf{51.7} & \textbf{40.9} &	\textbf{36.1}\\
\bottomrule
\end{tabular}}
\label{tab:sequential_vs_multi_task}
\vspace{-5mm}
\end{table}

\begin{figure*}[t]
\centering
\includegraphics[width=1.0\linewidth]{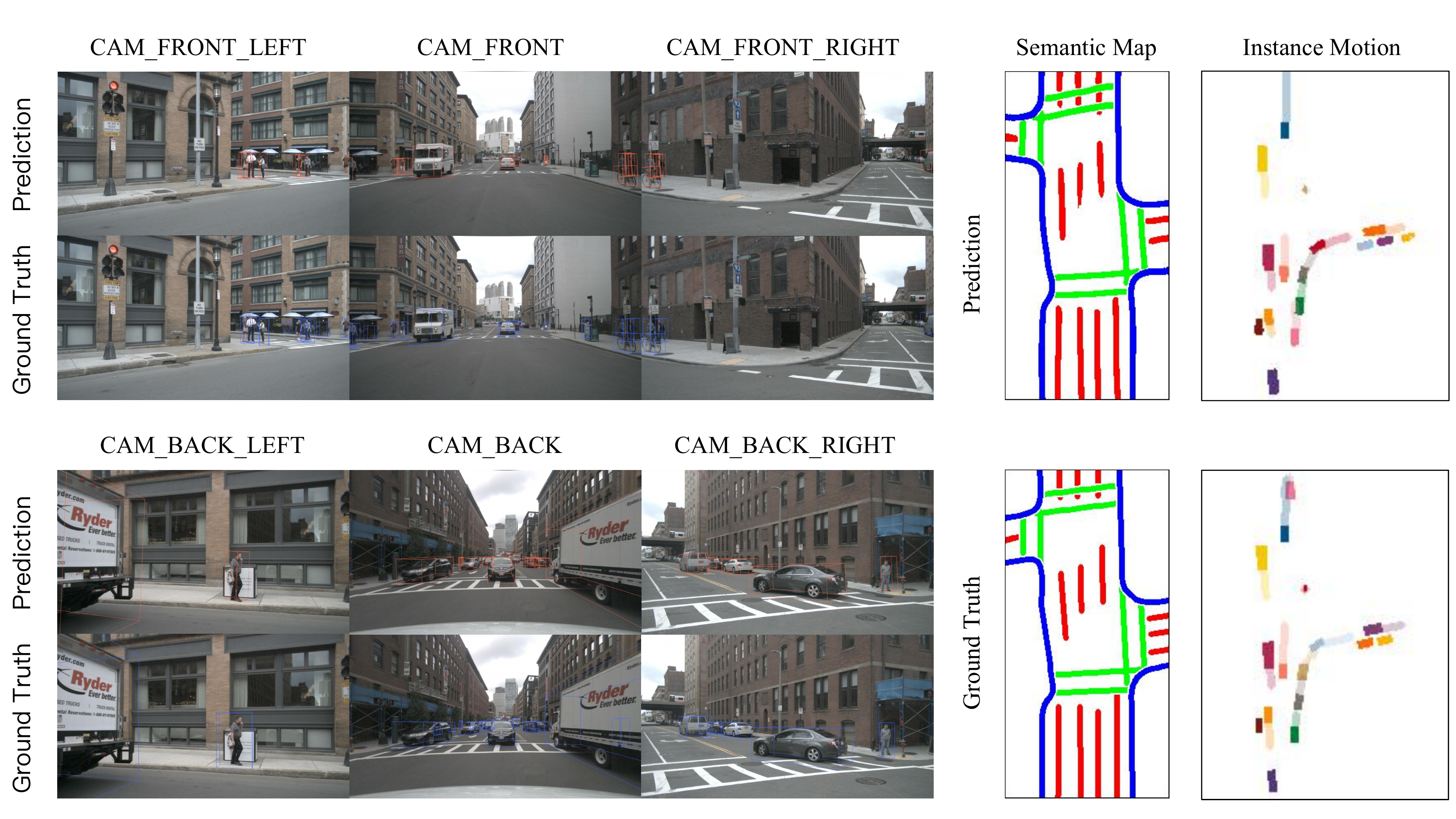}
\caption{The qualitative results for 3D object detection, semantic map construction, and instance motion prediction. For the visualization of semantic maps, we represent lane dividers, pedestrian crossings, and lane boundaries with red, green, and blue colors. For the visualization of motions, the future trajectories of road agents are shown with transparent paths. Best viewed in color. 
}
\label{fig:val_visualize}
\vspace{-5mm}
\end{figure*}

\noindent\textbf{Multi-task learning.} To analyze the influence of multi-task learning for each task, we train multiple single-task variants of BEVerse and summarize the experimental results in Table~\ref{tab:ablation_multi_task}. We can observe that BEVerse underperforms the single-task models for 3D object detection and semantic map construction. Since these two tasks focus, respectively, on the dynamic and static elements of the environment, the performance drop is understandable with the shared capacity. 
However, the joint reasoning of detection and semantic map can greatly improve the performance of motion prediction. It can be explained that the consideration of detection can improve the discriminative ability for objects and the information of semantic maps can provide priors about the roads. 
Also, we can find that the multi-task framework favors in reducing the network parameters and improving efficiency. 

\noindent\textbf{Data-augmentation strategies.} As shown in Table~\ref{tab:ablation_augmentation}, both the image-view and BEV augmentation techniques are crucial for the performances of all three tasks, especially for 3D object detection and motion prediction. With the enhancement of temporal information, BEVerse is more likely to overfit to limited training samples and requires stronger regularization for better performance. Also, applying the BEV augmentation alone can hurt the learning of semantic map construction, possibly because the traffic lines usually follow specific directions.

\subsection{Main Results}

\noindent\textbf{3D object detection.}
We compare BEVerse and previous methods for 3D object detection on the nuScenes detection benchmark. As shown in Table~\ref{tab:nus_det_val}, BEVerse-Tiny can achieve 46.6\% NDS and 32.1\% mAP, significantly outperforming BEVDet-Tiny with the same input size. BEVerse-Small further achieves the new state-of-the-art performance of 49.5\% NDS for 3D object detection, even with the multi-task framework. Also, the test-time augmentation can continue to boost the performance. 
The result on the test set is shown in Table~\ref{tab:nus_det_test}. BEVerse with 53.1\% NDS outperforms all existing published methods, without using external depth data for pre-training. 

\noindent\textbf{Semantic map construction.} In Table~\ref{tab:nus_map_val}, we benchmark the performance of BEVerse for semantic map construction. 
We can observe that 
BEVerse-Tiny already obtains the mIoU of 48.7 and outperforms existing methods.
Furthermore,
BEVerse-Small achieves 51.7 mIoU, which is 7.1 points higher than the previous best method~\cite{bevsegformer}.

\noindent\textbf{Motion prediction.}
The quantitative comparison for motion prediction between BEVerse and existing methods is summarized in Table~\ref{tab:nus_motion_val}. Considering the long ranges, BEVerse-Tiny achieves 38.7 IoU and 33.3 VPQ and significantly outperforms the only published method FIERY~\cite{fiery}. The performance is further improved with larger image sizes and backbones.

\noindent\textbf{Joint perception and prediction.} To demonstrate the superiority of our multi-task framework, we compare BEVerse and the sequential paradigm in Table~\ref{tab:sequential_vs_multi_task}. The sequential framework simply stacks three state-of-the-art methods for 3D object detection, semantic map construction, and motion prediction. We can observe that the proposed BEVerse not only outperforms in all three tasks, but operates much faster.
We demonstrate that the multi-task framework BEVerse provides a better trade-off with shared features and joint reasoning. 

\subsection{Qualitative Results}

As shown in Figure~\ref{fig:val_visualize}, we visualize the multi-task predictions of BEVerse-Small on the nuScenes~\cite{nuscenes} validation set. BEVerse can generate highly-accurate 3D bounding boxes for vehicles, pedestrians, cyclist and other objects. Also, the predicted semantic maps and future trajectories of vehicles are very close to the ground truth.
However, some errors in depth prediction can be observed for both faraway objects and traffic lines.

\section{Conclusion}

In this paper, we present BEVerse, the first unified framework for 3D perception and prediction based on multi-camera systems. Unlike existing methods for single tasks, BEVerse generates 4D BEV representations from multi-camera videos and jointly reasons about 3D object detection, semantic map construction, and motion prediction. With comprehensive experiments on the nuScenes dataset, we show that the multi-task paradigm of BEVerse can achieve the state-of-the-art performance on three tasks and is more efficient than the sequential paradigm. We also show that the joint learning of perception tasks can implicitly improve the motion prediction. 


{\small
\bibliographystyle{ieee_fullname}
\bibliography{egbib}

\begin{thebibliography}{10}\itemsep=-1pt

\bibitem{stretchbev}
Adil~Kaan Akan and Fatma G{\"u}ney.
\newblock Stretchbev: Stretching future instance prediction spatially and
  temporally.
\newblock {\em arXiv preprint arXiv:2203.13641}, 2022.

\bibitem{M3D-RPN}
Garrick Brazil and Xiaoming Liu.
\newblock M3d-rpn: Monocular 3d region proposal network for object detection.
\newblock In {\em ICCV}, 2019.

\bibitem{nuscenes}
Holger Caesar, Varun Bankiti, Alex~H Lang, Sourabh Vora, Venice~Erin Liong,
  Qiang Xu, Anush Krishnan, Yu Pan, Giancarlo Baldan, and Oscar Beijbom.
\newblock nuscenes: A multimodal dataset for autonomous driving.
\newblock In {\em CVPR}, 2020.

\bibitem{detr}
Nicolas Carion, Francisco Massa, Gabriel Synnaeve, Nicolas Usunier, Alexander
  Kirillov, and Sergey Zagoruyko.
\newblock End-to-end object detection with transformers.
\newblock In {\em ECCV}, 2020.

\bibitem{mono3d}
Xiaozhi Chen, Kaustav Kundu, Ziyu Zhang, Huimin Ma, Sanja Fidler, and Raquel
  Urtasun.
\newblock Monocular 3d object detection for autonomous driving.
\newblock In {\em CVPR}, 2016.

\bibitem{gradnorm}
Zhao Chen, Vijay Badrinarayanan, Chen-Yu Lee, and Andrew Rabinovich.
\newblock Gradnorm: Gradient normalization for adaptive loss balancing in deep
  multitask networks.
\newblock In {\em ICML}. PMLR, 2018.

\bibitem{gradient_sign_dropout}
Zhao Chen, Jiquan Ngiam, Yanping Huang, Thang Luong, Henrik Kretzschmar, Yuning
  Chai, and Dragomir Anguelov.
\newblock Just pick a sign: Optimizing deep multitask models with gradient sign
  dropout.
\newblock {\em NeurIPS}, 2020.

\bibitem{review-motion-planning}
Laurene Claussmann, Marc Revilloud, Dominique Gruyer, and S{\'e}bastien Glaser.
\newblock A review of motion planning for highway autonomous driving.
\newblock {\em IEEE Transactions on Intelligent Transportation Systems},
  21(5):1826--1848, 2019.

\bibitem{uncertainty-motion}
Nemanja Djuric, Vladan Radosavljevic, Henggang Cui, Thi Nguyen, Fang-Chieh
  Chou, Tsung-Han Lin, Nitin Singh, and Jeff Schneider.
\newblock Uncertainty-aware short-term motion prediction of traffic actors for
  autonomous driving.
\newblock In {\em WACV}, 2020.

\bibitem{SLAM_2017_iros}
Renaud Dub{\'e}, Abel Gawel, Hannes Sommer, Juan Nieto, Roland Siegwart, and
  Cesar Cadena.
\newblock An online multi-robot slam system for 3d lidars.
\newblock In {\em IROS}. IEEE, 2017.

\bibitem{tpnet}
Liangji Fang, Qinhong Jiang, Jianping Shi, and Bolei Zhou.
\newblock Tpnet: Trajectory proposal network for motion prediction.
\newblock In {\em CVPR}, 2020.

\bibitem{nddr-cnn}
Yuan Gao, Jiayi Ma, Mingbo Zhao, Wei Liu, and Alan~L Yuille.
\newblock Nddr-cnn: Layerwise feature fusing in multi-task cnns by neural
  discriminative dimensionality reduction.
\newblock In {\em CVPR}, 2019.

\bibitem{kitti}
Andreas Geiger, Philip Lenz, and Raquel Urtasun.
\newblock Are we ready for autonomous driving? the kitti vision benchmark
  suite.
\newblock In {\em CVPR}, 2012.

\bibitem{denseTNT}
Junru Gu, Chen Sun, and Hang Zhao.
\newblock Densetnt: End-to-end trajectory prediction from dense goal sets.
\newblock In {\em ICCV}, 2021.

\bibitem{ddad}
Vitor Guizilini, Rares Ambrus, Sudeep Pillai, Allan Raventos, and Adrien
  Gaidon.
\newblock 3d packing for self-supervised monocular depth estimation.
\newblock In {\em CVPR}, 2020.

\bibitem{learning_to_branch_mtl}
Pengsheng Guo, Chen-Yu Lee, and Daniel Ulbricht.
\newblock Learning to branch for multi-task learning.
\newblock In {\em ICML}. PMLR, 2020.

\bibitem{resnet}
Kaiming He, Xiangyu Zhang, Shaoqing Ren, and Jian Sun.
\newblock Deep residual learning for image recognition.
\newblock In {\em CVPR}, 2016.

\bibitem{planning_RL}
Carl-Johan Hoel, Katherine Driggs-Campbell, Krister Wolff, Leo Laine, and
  Mykel~J Kochenderfer.
\newblock Combining planning and deep reinforcement learning in tactical
  decision making for autonomous driving.
\newblock {\em IEEE transactions on intelligent vehicles}, 5(2):294--305, 2019.

\bibitem{fiery}
Anthony Hu, Zak Murez, Nikhil Mohan, Sof{\'\i}a Dudas, Jeffrey Hawke, Vijay
  Badrinarayanan, Roberto Cipolla, and Alex Kendall.
\newblock Fiery: Future instance prediction in bird's-eye view from surround
  monocular cameras.
\newblock In {\em ICCV}, 2021.

\bibitem{bevdet}
Junjie Huang, Guan Huang, Zheng Zhu, and Dalong Du.
\newblock Bevdet: High-performance multi-camera 3d object detection in
  bird-eye-view.
\newblock {\em arXiv preprint arXiv:2112.11790}, 2021.

\bibitem{batchnorm}
Sergey Ioffe and Christian Szegedy.
\newblock Batch normalization: Accelerating deep network training by reducing
  internal covariate shift.
\newblock {\em arXiv preprint arXiv:1502.03167}, 2015.

\bibitem{predict_parsing_motion}
Xiaojie Jin, Huaxin Xiao, Xiaohui Shen, Jimei Yang, Zhe Lin, Yunpeng Chen,
  Zequn Jie, Jiashi Feng, and Shuicheng Yan.
\newblock Predicting scene parsing and motion dynamics in the future.
\newblock {\em NeurIPS}, 2017.

\bibitem{what_uncertainty}
Alex Kendall and Yarin Gal.
\newblock What uncertainties do we need in bayesian deep learning for computer
  vision?
\newblock In {\em NeurIPS}, 2017.

\bibitem{krizhevsky2012imagenet}
Alex Krizhevsky, Ilya Sutskever, and Geoffrey~E Hinton.
\newblock Imagenet classification with deep convolutional neural networks.
\newblock In {\em NIPS}, pages 1097--1105, 2012.

\bibitem{pointpillars}
Alex~H Lang, Sourabh Vora, Holger Caesar, Lubing Zhou, Jiong Yang, and Oscar
  Beijbom.
\newblock Pointpillars: Fast encoders for object detection from point clouds.
\newblock In {\em CVPR}, 2019.

\bibitem{robust_pose_graph_EM}
Gim~Hee Lee, Friedrich Fraundorfer, and Marc Pollefeys.
\newblock Robust pose-graph loop-closures with expectation-maximization.
\newblock In {\em IROS}. IEEE, 2013.

\bibitem{DESIRE}
Namhoon Lee, Wongun Choi, Paul Vernaza, Christopher~B Choy, Philip~HS Torr, and
  Manmohan Chandraker.
\newblock Desire: Distant future prediction in dynamic scenes with interacting
  agents.
\newblock In {\em CVPR}, 2017.

\bibitem{GS3D}
Buyu Li, Wanli Ouyang, Lu Sheng, Xingyu Zeng, and Xiaogang Wang.
\newblock Gs3d: An efficient 3d object detection framework for autonomous
  driving.
\newblock In {\em CVPR}, 2019.

\bibitem{hdmapnet}
Qi Li, Yue Wang, Yilun Wang, and Hang Zhao.
\newblock Hdmapnet: A local semantic map learning and evaluation framework.
\newblock {\em arXiv preprint arXiv:2107.06307}, 2021.

\bibitem{PNPNet}
Ming Liang, Bin Yang, Wenyuan Zeng, Yun Chen, Rui Hu, Sergio Casas, and Raquel
  Urtasun.
\newblock Pnpnet: End-to-end perception and prediction with tracking in the
  loop.
\newblock In {\em CVPR}, 2020.

\bibitem{FQNet}
Lijie Liu, Jiwen Lu, Chunjing Xu, Qi Tian, and Jie Zhou.
\newblock Deep fitting degree scoring network for monocular 3d object
  detection.
\newblock In {\em CVPR}, 2019.

\bibitem{PETR}
Yingfei Liu, Tiancai Wang, Xiangyu Zhang, and Jian Sun.
\newblock Petr: Position embedding transformation for multi-view 3d object
  detection.
\newblock {\em arXiv preprint arXiv:2203.05625}, 2022.

\bibitem{swin_transformer}
Ze Liu, Yutong Lin, Yue Cao, Han Hu, Yixuan Wei, Zheng Zhang, Stephen Lin, and
  Baining Guo.
\newblock Swin transformer: Hierarchical vision transformer using shifted
  windows.
\newblock In {\em ICCV}, 2021.

\bibitem{autoshape}
Zongdai Liu, Dingfu Zhou, Feixiang Lu, Jin Fang, and Liangjun Zhang.
\newblock Autoshape: Real-time shape-aware monocular 3d object detection.
\newblock In {\em ICCV}, 2021.

\bibitem{multilinear_relation}
Mingsheng Long, Zhangjie Cao, Jianmin Wang, and Philip~S Yu.
\newblock Learning multiple tasks with multilinear relationship networks.
\newblock {\em NeurIPS}, 2017.

\bibitem{adamw}
Ilya Loshchilov and Frank Hutter.
\newblock Decoupled weight decay regularization.
\newblock In {\em ICLR}, 2019.

\bibitem{GUPNet}
Yan Lu, Xinzhu Ma, Lei Yang, Tianzhu Zhang, Yating Liu, Qi Chu, Junjie Yan, and
  Wanli Ouyang.
\newblock Geometry uncertainty projection network for monocular 3d object
  detection.
\newblock In {\em ICCV}, 2021.

\bibitem{FAF}
Wenjie Luo, Bin Yang, and Raquel Urtasun.
\newblock Fast and furious: Real time end-to-end 3d detection, tracking and
  motion forecasting with a single convolutional net.
\newblock In {\em CVPR}, 2018.

\bibitem{porca_planning}
Yuanfu Luo, Panpan Cai, Aniket Bera, David Hsu, Wee~Sun Lee, and Dinesh
  Manocha.
\newblock Porca: Modeling and planning for autonomous driving among many
  pedestrians.
\newblock {\em IEEE Robotics and Automation Letters}, 3(4):3418--3425, 2018.

\bibitem{monodle}
Xinzhu Ma, Yinmin Zhang, Dan Xu, Dongzhan Zhou, Shuai Yi, Haojie Li, and Wanli
  Ouyang.
\newblock Delving into localization errors for monocular 3d object detection.
\newblock In {\em CVPR}, 2021.

\bibitem{voxelnet}
Daniel Maturana and Sebastian Scherer.
\newblock Voxnet: A 3d convolutional neural network for real-time object
  recognition.
\newblock In {\em IROS}, 2015.

\bibitem{motion-planning-ICRA-2011}
Matthew McNaughton, Chris Urmson, John~M Dolan, and Jin-Woo Lee.
\newblock Motion planning for autonomous driving with a conformal
  spatiotemporal lattice.
\newblock In {\em ICRA}. IEEE, 2011.

\bibitem{icp_pose_graph_SLAM}
Ellon Mendes, Pierrick Koch, and Simon Lacroix.
\newblock Icp-based pose-graph slam.
\newblock In {\em IEEE International Symposium on Safety, Security, and Rescue
  Robotics (SSRR)}. IEEE, 2016.

\bibitem{cross-stitch}
Ishan Misra, Abhinav Shrivastava, Abhinav Gupta, and Martial Hebert.
\newblock Cross-stitch networks for multi-task learning.
\newblock In {\em CVPR}, 2016.

\bibitem{deep3Dbox}
Arsalan Mousavian, Dragomir Anguelov, John Flynn, and Jana Kosecka.
\newblock 3d bounding box estimation using deep learning and geometry.
\newblock In {\em CVPR}, 2017.

\bibitem{centerfusion}
Ramin Nabati and Hairong Qi.
\newblock Centerfusion: Center-based radar and camera fusion for 3d object
  detection.
\newblock In {\em WACV}, 2021.

\bibitem{VPN}
Bowen Pan, Jiankai Sun, Ho~Yin~Tiga Leung, Alex Andonian, and Bolei Zhou.
\newblock Cross-view semantic segmentation for sensing surroundings.
\newblock {\em IEEE Robotics and Automation Letters}, 2020.

\bibitem{dd3d}
Dennis Park, Rares Ambrus, Vitor Guizilini, Jie Li, and Adrien Gaidon.
\newblock Is pseudo-lidar needed for monocular 3d object detection?
\newblock In {\em ICCV}, 2021.

\bibitem{bevsegformer}
Lang Peng, Zhirong Chen, Zhangjie Fu, Pengpeng Liang, and Erkang Cheng.
\newblock Bevsegformer: Bird's eye view semantic segmentation from arbitrary
  camera rigs.
\newblock {\em arXiv preprint arXiv:2203.04050}, 2022.

\bibitem{LSS}
Jonah Philion and Sanja Fidler.
\newblock Lift, splat, shoot: Encoding images from arbitrary camera rigs by
  implicitly unprojecting to 3d.
\newblock In {\em ECCV}, 2020.

\bibitem{JointLP2}
John Phillips, Julieta Martinez, Ioan~Andrei B{\^a}rsan, Sergio Casas, Abbas
  Sadat, and Raquel Urtasun.
\newblock Deep multi-task learning for joint localization, perception, and
  prediction.
\newblock In {\em CVPR}, 2021.

\bibitem{safety-aware-motion}
Xuanchi Ren, Tao Yang, Li~Erran Li, Alexandre Alahi, and Qifeng Chen.
\newblock Safety-aware motion prediction with unseen vehicles for autonomous
  driving.
\newblock In {\em ICCV}, 2021.

\bibitem{overview_mtl_2017}
Sebastian Ruder.
\newblock An overview of multi-task learning in deep neural networks.
\newblock {\em arXiv preprint arXiv:1706.05098}, 2017.

\bibitem{latent_mtl}
Sebastian Ruder, Joachim Bingel, Isabelle Augenstein, and Anders S{\o}gaard.
\newblock Latent multi-task architecture learning.
\newblock In {\em AAAI}, 2019.

\bibitem{lio-sam}
Tixiao Shan, Brendan Englot, Drew Meyers, Wei Wang, Carlo Ratti, and Daniela
  Rus.
\newblock Lio-sam: Tightly-coupled lidar inertial odometry via smoothing and
  mapping.
\newblock In {\em IROS}. IEEE, 2020.

\bibitem{disentangling}
Andrea Simonelli, Samuel~Rota Bulo, Lorenzo Porzi, Manuel L{\'o}pez-Antequera,
  and Peter Kontschieder.
\newblock Disentangling monocular 3d object detection.
\newblock In {\em CVPR}, 2019.

\bibitem{waymo}
Pei Sun, Henrik Kretzschmar, Xerxes Dotiwalla, Aurelien Chouard, Vijaysai
  Patnaik, Paul Tsui, James Guo, Yin Zhou, Yuning Chai, Benjamin Caine, et~al.
\newblock Scalability in perception for autonomous driving: Waymo open dataset.
\newblock In {\em CVPR}, 2020.

\bibitem{fcos}
Zhi Tian, Chunhua Shen, Hao Chen, and Tong He.
\newblock Fcos: Fully convolutional one-stage object detection.
\newblock In {\em CVPR}, 2019.

\bibitem{transformer}
Ashish Vaswani, Noam Shazeer, Niki Parmar, Jakob Uszkoreit, Llion Jones,
  Aidan~N Gomez, {\L}ukasz Kaiser, and Illia Polosukhin.
\newblock Attention is all you need.
\newblock In {\em NeurIPS}, 2017.

\bibitem{patch_to_the_future}
Jacob Walker, Abhinav Gupta, and Martial Hebert.
\newblock Patch to the future: Unsupervised visual prediction.
\newblock In {\em CVPR}, 2014.

\bibitem{PGD}
Tai Wang, ZHU Xinge, Jiangmiao Pang, and Dahua Lin.
\newblock Probabilistic and geometric depth: Detecting objects in perspective.
\newblock In {\em CORL}, 2022.

\bibitem{fcos3d}
Tai Wang, Xinge Zhu, Jiangmiao Pang, and Dahua Lin.
\newblock Fcos3d: Fully convolutional one-stage monocular 3d object detection.
\newblock In {\em ICCV}, 2021.

\bibitem{pseudo-lidar}
Yan Wang, Wei-Lun Chao, Divyansh Garg, Bharath Hariharan, Mark Campbell, and
  Kilian~Q Weinberger.
\newblock Pseudo-lidar from visual depth estimation: Bridging the gap in 3d
  object detection for autonomous driving.
\newblock In {\em CVPR}, 2019.

\bibitem{detr3d}
Yue Wang, Vitor Guizilini, Tianyuan Zhang, Yilun Wang, Hang Zhao, , and
  Justin~M. Solomon.
\newblock Detr3d: 3d object detection from multi-view images via 3d-to-2d
  queries.
\newblock In {\em CoRL}, 2021.

\bibitem{motionnet}
Pengxiang Wu, Siheng Chen, and Dimitris~N Metaxas.
\newblock Motionnet: Joint perception and motion prediction for autonomous
  driving based on bird's eye view maps.
\newblock In {\em CVPR}, 2020.

\bibitem{second}
Yan Yan, Yuxing Mao, and Bo Li.
\newblock Second: Sparsely embedded convolutional detection.
\newblock {\em Sensors}, 2018.

\bibitem{robust_pose_graph_for_SLAM}
Sheng Yang, Xiaoling Zhu, Xing Nian, Lu Feng, Xiaozhi Qu, and Teng Ma.
\newblock A robust pose graph approach for city scale lidar mapping.
\newblock In {\em IROS}. IEEE, 2018.

\bibitem{centertrack}
Tianwei Yin, Xingyi Zhou, and Philipp Kr{\"a}henb{\"u}hl.
\newblock Center-based 3d object detection and tracking.
\newblock {\em CVPR}, 2021.

\bibitem{monopair}
Chen Yongjian, Tai Lei, Sun Kai, and Li Mingyang.
\newblock Monopair: Monocular 3d object detection using pairwise spatial
  relationships.
\newblock In {\em CVPR}, 2020.

\bibitem{gradient_surgery}
Tianhe Yu, Saurabh Kumar, Abhishek Gupta, Sergey Levine, Karol Hausman, and
  Chelsea Finn.
\newblock Gradient surgery for multi-task learning.
\newblock {\em NeurIPS}, 2020.

\bibitem{monoflex}
Yunpeng Zhang, Jiwen Lu, and Jie Zhou.
\newblock Objects are different: Flexible monocular 3d object detection.
\newblock In {\em CVPR}, 2021.

\bibitem{overview_mtl_2018}
Yu Zhang and Qiang Yang.
\newblock An overview of multi-task learning.
\newblock {\em National Science Review}, 5(1):30--43, 2018.

\bibitem{tnt}
Hang Zhao, Jiyang Gao, Tian Lan, Chen Sun, Benjamin Sapp, Balakrishnan
  Varadarajan, Yue Shen, Yi Shen, Yuning Chai, Cordelia Schmid, et~al.
\newblock Tnt: Target-driven trajectory prediction.
\newblock {\em arXiv preprint arXiv:2008.08294}, 2020.

\bibitem{centernet}
Xingyi Zhou, Dequan Wang, and Philipp Kr{\"a}henb{\"u}hl.
\newblock Objects as points.
\newblock {\em arXiv preprint arXiv:1904.07850}, 2019.

\bibitem{cbgs}
Benjin Zhu, Zhengkai Jiang, Xiangxin Zhou, Zeming Li, and Gang Yu.
\newblock Class-balanced grouping and sampling for point cloud 3d object
  detection.
\newblock {\em arXiv preprint arXiv:1908.09492}, 2019.

\end{thebibliography}
}

\end{document}